\documentclass[runningheads]{llncs}
\usepackage{graphicx,import} 
\usepackage{transparent}
\usepackage{booktabs}
\usepackage{mathrsfs}
\usepackage[inline]{enumitem}
\usepackage{circuitikz}
\usetikzlibrary{shapes,shadows,arrows,automata}
\usepackage{todonotes}
\usepackage{algorithm}
\usepackage{algpseudocode}
\usepackage{subfig}
\usepackage{amsmath}
\usepackage{amsfonts}
\begin{document}
\title{Failure-Tolerant Connectivity Maintenance for Robot Swarms}
%
%
\author{Vivek Shankar Varadharajan\inst{1} \and 
Bram Adams\inst{1} \and 
Giovanni Beltrame\inst{1} } 
\authorrunning{V. S. Varadharajan et al.}
%
\institute{Polytechnique Montréal, Montréal, QC H3T 1J4, Canada}
\maketitle              
\begin{abstract}
  Connectivity maintenance plays a key role in achieving a desired global
  behaviour among a swarm of robots. However, connectivity maintenance in realistic environments is hampered by lack of computation resources, low
  communication bandwidth, robot failures and unstable links. In this
  paper, we propose a novel decentralized connectivity-preserving algorithm
  that can be deployed on top of other behaviours to enforce connectivity
  constraints. The algorithm takes a set of targets to be reached while
  keeping a minimum number of redundant links between robots, with the goal of
  guaranteeing bandwidth and reliability. Robots then incrementally build and maintain a communication backbone with the specified
  number of links. We empirically study the performance of the algorithm,
  analyzing its time to convergence, as well as robustness to faults injected into the backbone robots. Our results statistically
  demonstrate the algorithm's ability to preserve the desired connectivity
  constraints and to reach the targets with up to 70 percent of individual robot failures in the
  communication backbone.

\end{abstract}


\section{Introduction}
Swarm robotics is a field of engineering that deals with relatively simple
physically agents to achieve a global behaviour that emerges as a result of
local interactions~\cite{sahin2005}. Swarm robotics has been widely
investigated in the last decade~\cite{Brambilla2013} for a number of different
applications, mainly due to its inherent benefits: robustness, scalablity, and
flexibility. With a large swarm, in general, the loss of a single agent does
not jeopardize the overall mission and a failed agent could be replaced with
another. Robotic swarms are deemed cost effective solutions when dealing with
large, spatially distributed tasks like
exploration~\cite{manjanna2018heterogeneous}, search and
rescue~\cite{Sampedro2018}, and area coverage~\cite{Giuggioli2018}.

In many of these applications the robots need communication between each other to coordinate. For
the information to propagate, the swarm needs to be \emph{connected},
i.e. there has to be a communication path between all the robots in the
swarm. The problem of \emph{maintaining connectivity} is widely discussed in
literature, with a number of different recent
approaches~\cite{panerati2018swarms,gasparri2017bounded,majcherczyk2018decentralized,panerati2018robust}. Some
of these approaches design control strategies to enforce algebraic
connectivity~\cite{fiedler1973} among a group of connected nodes~\cite{ghedini2016improving,minelli2018stop,sabattini2013decentralized,gasparri2017bounded}, while others
address the connectivity problem by enforcing virtual forces on a pre-existent
structure~\cite{krupke2015distributed,soleymani2015distributed,schuresko2012distributed}.

During a real world mission robots can fail for a number of reasons
(environmental factors, wear and tear, etc) and break connectivity,
compromising the mission. In addition, maintaining connectivity is likely not
the only requirement for mission success. Consider the Fukushima accident in
2011: robots were used to inspect the collapsed nuclear power plant (with a
video feed), and they were subject to extremely high failure rates due to
radiation.
The work in~\cite{panerati2018robust} considers a robustness factor into the
designed control law to tackle robot failures while enforcing
connectivity. The convergence in~\cite{panerati2018robust} is slow due to the
computation of the Fiedler vector (an algebraic connectiviy measure) using a
power iteration method, which requires multiple information exchanges
throughout the entire swarm.  In this paper, we propose a decentralized, failure-tolerant
connectivity maintenance approach that can be added to existing control
algorithms. Our approach is lightweight in both communication and computation
requirements, freeing resources to achieve the mission goals.

In practice, we progressively and dynamically use the robots in the swarm to
form a communication backbone from a root to a set of targets. We set the
number of links between the robots as a configurable redundancy factor. The
contributions of the work can be summarized as:
\begin{enumerate}
\item formalization of a chain-based backbone algorithm, that progressively places the
  robots towards the targets with a configurable number of links;
\item study of the algorithm performance using a physics-based simulator;
\item analysis on the performance of the algorithm with simulated robot failures.
\end{enumerate}

The paper is structured as follows: a brief summary of the related work in
literature is given in Section \ref{sec:literature}; we describe the
mathematical model (kinematic and communication) in Section~\ref{sec:model}; our proposed algorithm is described in Section~\ref{sec:approach}; Section~\ref{sec:experiments} provides
experimental results and analysis; finally, Section~\ref{sec:conclusions}
draws some concluding remarks.

\section{Related work\label{sec:literature}}
The problem of maintaining connectivity in a multi-robot system was addressed
in several ways in the literature. One approach is to design reactive control
laws while imposing connectivity as a constraint. For example, in
\cite{ji2007distributed} two control laws were developed for rendezvous and
formation tasks, imposing an initially connected configuration to be an
invariant set, effectively preserving connectivity. A similar approach was
implemented in \cite{su2010rendezvous}, achieving rendezvous among a group of
agents and preserving an initial connected
condition using a potential based controller.
This class of approaches relies on global coordination, reducing their overall
scalability, and making them more appropriate for small groups of robots.

Other works are more explicit, and use control laws that maximize algebraic
connectivity (i.e. the second eigenvalue of the Laplacian of the robots'
connectivity graph). One example is \cite{sabattini2011decentralized}, which
describes a method for the distributed estimation of algebraic connectivity
using a power iteration. Sabattini et al.~\cite{sabattini2011decentralized}
uses this estimation to drive agents in a way that maximizes connectivity. Kim et
al.~\cite{kim2005maximizing} proposes another approach that depends on solving
an optimization problem on the relative locations of robots, maximizing the
second eigenvalue of the Laplacian. Another example~\cite{de2006decentralized}
uses a distributed approach to calculate the Fiedler vector (the eigenvector
corresponding to the second eigenvalue of the Laplacian), then estimate other
relevant elements of Laplacian for each agent in a decentralized manner, and
finally use these to derive a gradient based control law that maximizes
connectivity. The biggest advantage of these methods is that they work fairly
well for any topology. However, the downside of using algebraic connectivity
is that the distributed estimation of the adjacency matrix that is needed to
compute the Laplacian requires multiple iterations of information flow through
the graph to converge to a reasonable value. The time consumed to compute the
algebraic connectivity makes it a very brittle estimate in case of noise, and
limits its applicability to real world missions.

Another class of methods enforce the desired connectivity among robots by
constructing a given communication topology, a spanning tree for
instance. Schuresko et al.~\cite{schuresko2009distributed} describes a robust
and mission-agnostic algorithm to generate a spanning tree. Aragues et
al.~\cite{aragues2014triggered} implements an area coverage mission while preserving
a minimum spanning tree among robots. The approach keeps connectivity with
minimal interference on the area coverage mission. However, this method requires a
specific initial condition that cannot always be guaranteed during a real
world deployment. Majcherczyk et al.~\cite{majcherczyk2018decentralized} treats
the problem of decentralized deployment of multiple robots to different target
locations while preserving connectivity. The algorithm
in~\cite{majcherczyk2018decentralized} defines different roles for robots,
such that when a target is specified, a branch of the robot network is
deployed and additional robots are supplied to build a structure reaching the
target while maintaining connectivity. Panerati et. al.~\cite{panerati2018swarms} proposed a hybrid methodology with a navigation controller enforcing connectivity and a global scheduler to provide the navigation controller with optimal policies. Despite the use of a global scheduler to support the navigation controller, the approach is relatively slow in comparison with our approach. 

Our work is comparable to ~\cite{majcherczyk2018decentralized}, as we
dynamically build structures to reach given targets. The main difference is
that we are able to specify the number of redundant links required for a
particular target, including fault-tolerance in our method. Moreover, our
approach aims at minimizing the computation and communication load
of each robot, allowing deployment of other behaviours on top of the
connectivity maintenance algorithm.

\section{Model\label{sec:model}}
\subsection{Communication model}
We assume our robots to have situated communication~\cite{Stoy2001}: senders broadcast
messages within a limited range R, and receivers within this range estimate the
relative position of the sender in their local coordinate frame.

Inter-agent communication can be modelled as an undirected graph
$\mathcal{G} = (\nu,\epsilon,A)$, with $\nu=\{r_1,..r_N\}$ the node set
representing the robots, and $\epsilon$ being the edge set representing
communication links. An edge between $r_i$ and $r_j$ exists in $\epsilon$,
if and only if, the distance $\lVert \mathbf{p}_i - \mathbf{p}_j \rVert
\leq R$, with $\mathbf{p}_i$ and $\mathbf{p}_j$ being the i and j node positions respectively, $R$ being the communication range.

Spectral graph theory~\cite{cvetkovic2004} offers methods to estimate the
algebraic connectivity of a given graph using the adjacency matrix $A$ and the
Laplacian $L$. The Laplacian of a graph can be estimated using the adjacency
matrix and the degree matrix $D$. An entry $a_{ij}$ in the adjacency matrix is $1$ if the
edge $(i,j) \in \epsilon$ exists, $0$ otherwise. The adjacency matrix of an
undirected graph is symmetric. The degree matrix $D$ is a diagonal matrix
denoting the number of edges to a node. The Laplacian $L$ is defined as $L=D-A$.
The algebraic connectivity or Fiedler value~\cite{fiedler1973} is defined as
being the smallest non-zero eigenvalue $\lambda_2$ of $L$. Notably, if
$\lambda_2 > 0$ the graph is connected. An undirected graph is said to be
strongly connected if there exists a path between any two nodes in
the graph. Later in the experimental section~\ref{sec:experiments}, we study the evolution of network topology using the algebraic connectivity.     

\subsection{Robot Kinematics}
Let the state of the robot i be its position $p_i \in \mathbb{R}^m$ and let its state at time t be $p_i(t)$. The state of the swarm at time $t$ be the vector
$P(t) = \{p_1(t), p_2(t), ... , p_N(t)\}$. We assume the kinematics of the
robots to be
\begin{equation}
    \label{equ:kmodel}
    \begin{bmatrix}
        \Dot{x_i}(t) \\
        \Dot{y_i}(t) \\
        \Dot{\Theta_i}(t)
    \end{bmatrix}
    =
    \begin{bmatrix}
        \cos{\theta_i(t)} & 0 \\
        \sin{\theta_i(t)} & 0 \\
        0                 & 1
    \end{bmatrix}
    \cdot 
    \begin{bmatrix}
        v_i(t) \\
        \theta_i(t)
    \end{bmatrix}
\end{equation}
where the velocity of the robot can be directly controlled. We consider
2-dimensional non-holonomic robots with differential drive, hence
$p_i(t) = {[x_i(t) , y_i(t) , \theta_i(t) ]}^\top$.

Our connectivity maintenance algorithm issues high level commands in terms of
desired velocity, so that when a robot joins a chain to reach a target, we
apply simple PID control on the linear and angular velocities in
Equation~\ref{equ:kmodel}.

\subsection{Objectives}
The objective of this work is to incrementally construct a communication
backbone using a minimal number of robots, considering the communication
requirements of the targets. Let, $\mathscr{T} =\{t_1,t_2, ...,t_x\}$ be the
set of targets to be reached, $\mathscr{L}_i = \{l_1, l_2, ..., l_y\}$ be the
set of chains in any given target $t_i$. A chain $l_i$ corresponds to the communication chain $i$ connecting a robot visiting a target $t_i$, to the swarm. $r^{i}_{j} =\{r_1, r_2, ..., r_z\}$ be the set of robot agents forming the chain $j$ to a target $i$. The global objective of the backbone
construction is:
\begin{equation}
    \label{equ:gopt}
    \min{ \sum_{i \in \forall \mathscr{T}}\sum_{ j \in \forall \mathscr{L}^{i} }  {\lvert r^{i}_{j} \rvert} }   
\end{equation}
subjected to,
\begin{equation}
    \label{equ:gopt_con1}
    nl(t_i, r_i) \geq Nt_i,  \forall t_i \in T  
\end{equation}
\begin{equation}
    \label{equ:gopt_con2}
     v(t_i, r_i) = 1, \forall t_i \in T
\end{equation}
\begin{equation}
    \label{equ:gopt_lambda2}
    \lambda_2(dt) > 0, \forall dt \in T_t 
\end{equation}
Equation~\ref{equ:gopt} minimizes the number of robots in each link of all the
targets to be reached, i.e., robots used to build the communication
chains. Let $r_i$ be the robot visiting the target i, $v(t_i, r_i)$ denotes
whether the robot $r_i$ reached the target $t_i$.
\begin{equation}
    v(t_i, r_i)=
    \begin{cases}
      1, & \text{if link}\ r_{i}\ \text{reached target }t_i \\
      0, & \text{otherwise}
    \end{cases}
\end{equation}
Equation~\ref{equ:gopt_con2} ensures all the targets in set $\mathscr{T}$ are
reached. Let $nl(t_i, r_i)$ be the number of links between the swarm and robot
$r_i$ currently visiting target i, and $nl(t_i)$ be number of links required
to target i. The inequality in Equation~\ref{equ:gopt_con1} ensures that there
exists at least $Nt_i$ links between the swarm and robot $r_i$.
Equation~\ref{equ:gopt_lambda2} ensures that the resulting graph of the robot network is always connected.

\begin{figure*}
\subfloat[]{\includegraphics[height=0.5\linewidth, width=0.5\linewidth]{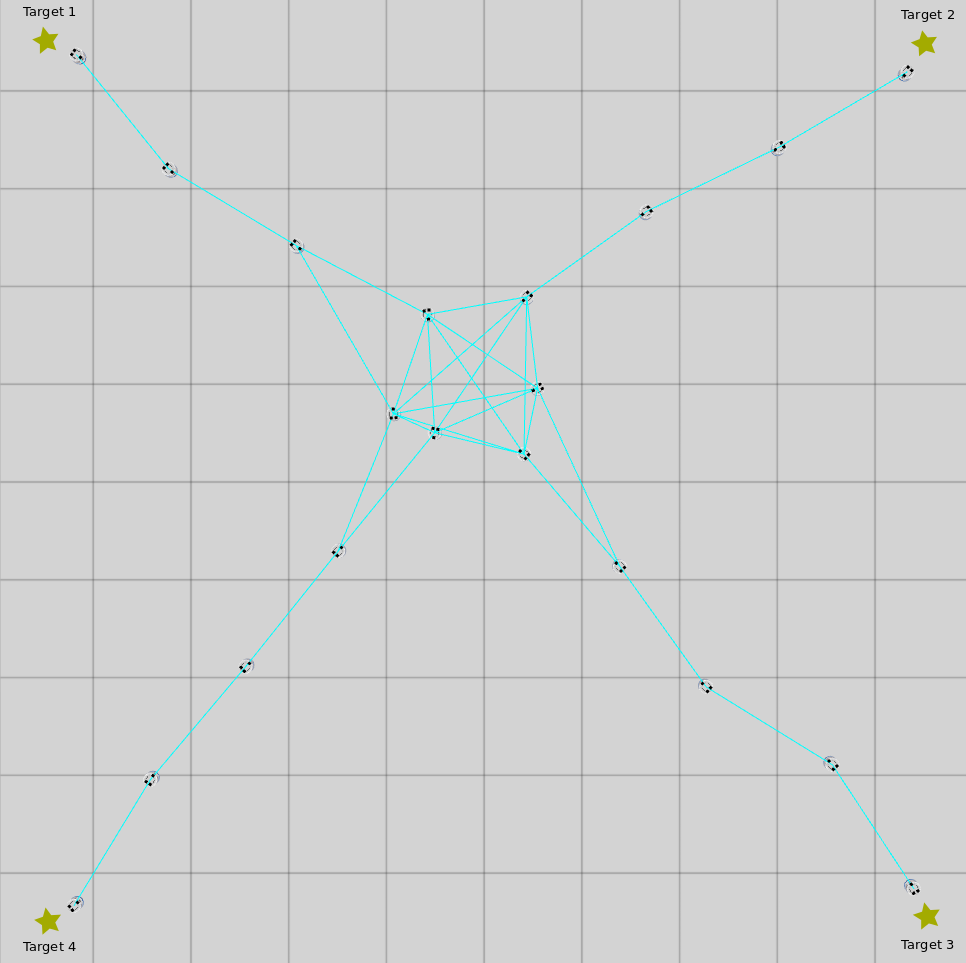}}\hfill
\subfloat[]{\includegraphics[height=0.5\linewidth, width=0.5\linewidth]{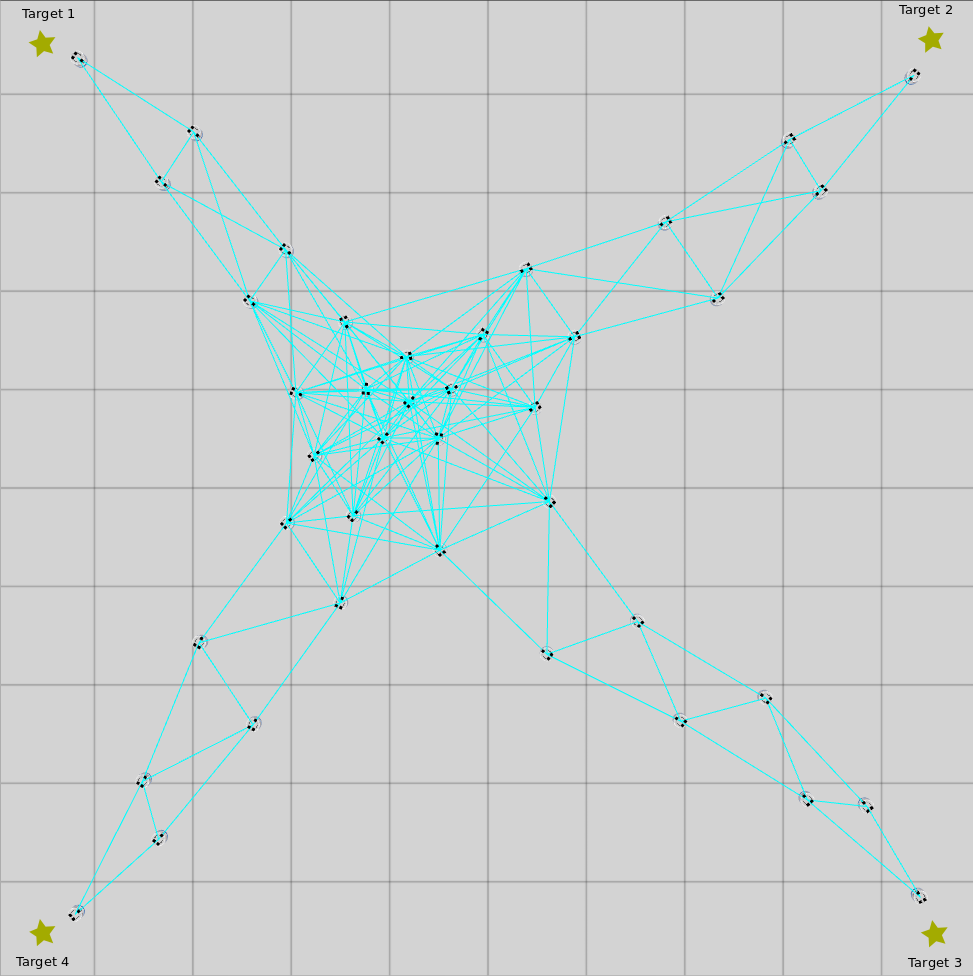}}\hfill\\
\subfloat[]{\includegraphics[height=0.5\linewidth, width=0.5\linewidth]{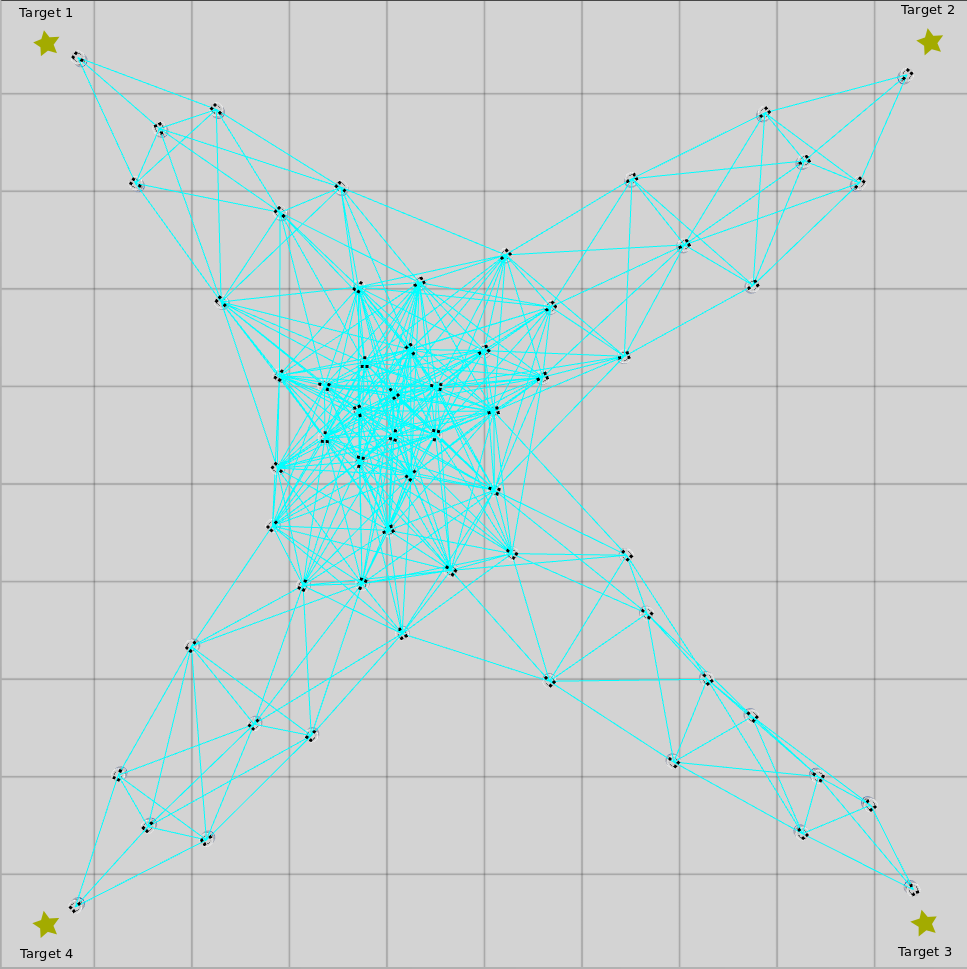}}\hfill
\subfloat[]{\includegraphics[height=0.5\linewidth, width=0.5\linewidth]{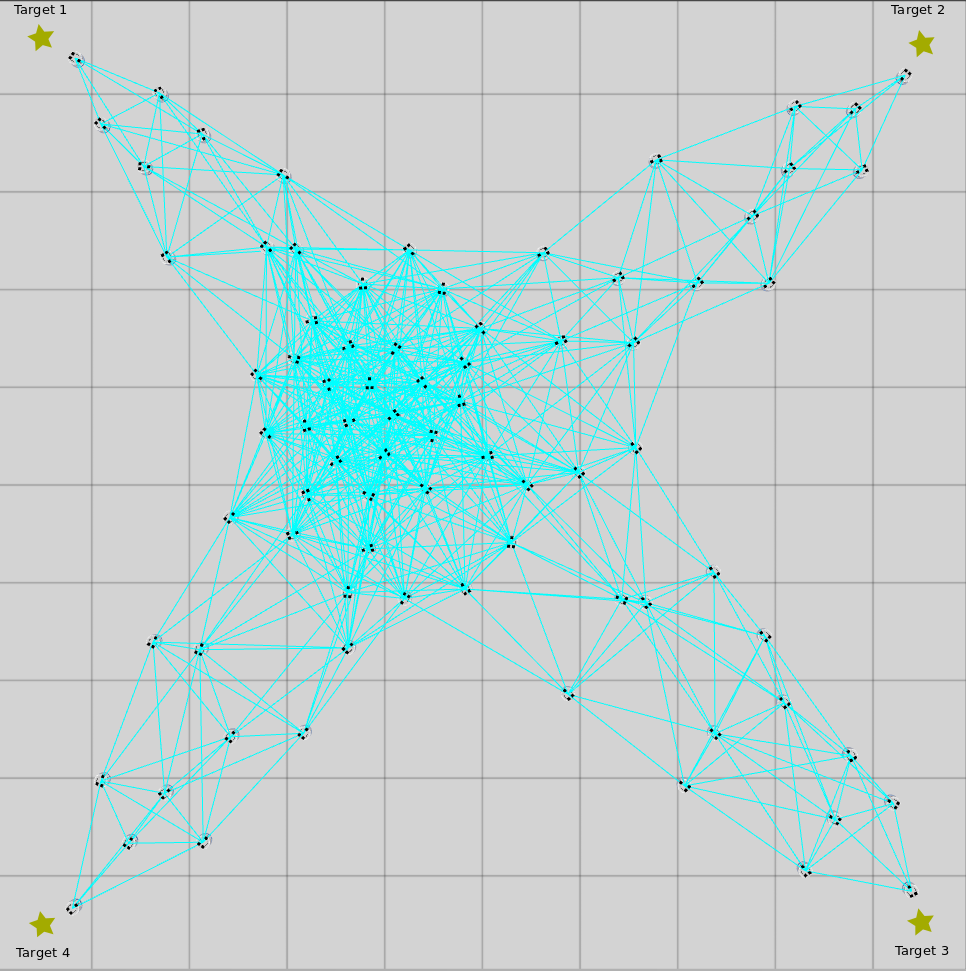}}\hfill
\caption{Screen capture of the simulation result: (a) 20 robots creating a single communication chain, (b) 40 robots creating 2 communication chains, (c) 60 robots creating 3 communication chains and (d) 80 robots creating 4 communication chains (the green inter-agent connections indicate the ability to communicate).}
\label{Fig:argos_sim}
\end{figure*}

\section{Approach\label{sec:approach}}
\subsection{Top-down specification}
In this work, the robots in the swarm are self organizing, with a global
behaviour emerging as a result of local interactions among robots. We assume that
the robots are randomly deployed, and we assume that the robot network is initially
connected. We believe this initial condition to be reasonable, since the
robots are deployed as a cluster from a deployment area in real world
scenarios. The desired global behavior of the swarm is to construct a tree
from a central reference robot to the robots visiting one or more target,
using minimal number of robots as in~\ref{equ:gopt}.

For this purpose, we build a communication chain for the robots visiting
each target $t_i$ from the target set $T$. We specify a target $t_i$ by its
position, orientation, and required number of links. We assume these
requirements to be variable: in a real world mission they might depend on what
is accomplished at the target. For instance, in an exploration mission the
targets could be landmarks from which photos or videos are required. A video
capture might require more bandwidth than the a photo
capture, resulting in a different number of links to achieve a desired bandwidth.

This algorithm assigns four types of roles to the robots in a swarm, namely:
\begin{enumerate*}
    \item root,
    \item free,
    \item networker, and,
    \item worker.
\end{enumerate*}
The root is assumed to be the center of the communication chains. Robots in
the swarm start to build a tree incrementally from the root robot. Worker
robots are assumed to be the robots visiting a target at a distant
location. The robots with the networker role maintain a certain distance from
their neighbors to secure a communication link. These robots use the control
law in Equation~\ref{equ:spring_Damp}. Free robots form a cluster around the
root, waiting for a networker or a worker to be selected to serve as a
relay. At first, the worker and root robots are selected. It would be ideal to
select a robot that is closest to the target as worker. We assume that robots
with worker roles are assigned to all targets in $T$ in advance, using a task
allocation algorithm for example~\cite{Lee2018}. A free
robot can either switch to be a networker or a worker depending on the
immediate need of the swarm.

Once the worker robots and the root robot are selected, the worker extends the
communication chain starting from the root. When the worker robot
determines it has reached a threshold distance $d_s$ from the root, the robot chooses a free
robot as a networker to act as relay to the root. Subsequently, when a
networker reaches a suitable distance, it selects a new free robot to serve as
a networker, and so on until the worker reaches the desired target.

\begin{figure}[!htb]
	\centering
	\resizebox{\linewidth}{!}{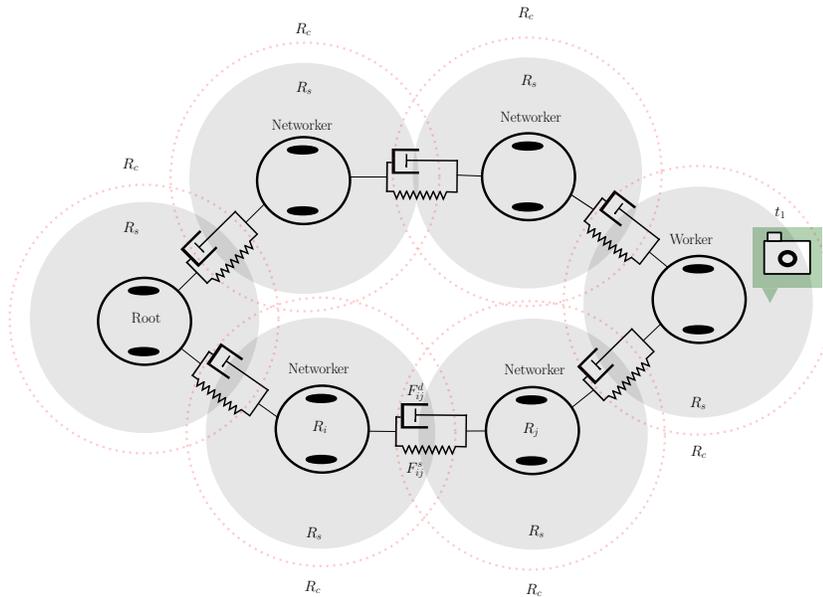}
	\caption{Spring and damper illustration of the interaction between the robots in a chain.}
	\label{fig:Springdamp}
\end{figure}


One way to model the interaction between the robots in a chain from a worker
to the root is by using the notion of virtual springs and
dampers~\cite{Wiech2018}, as shown in Figure~\ref{fig:Springdamp}. Each robot
exerts a virtual force ($FC_{ij}$) on the other to stay within the safe
communication distance ($d_s$). The exerted force is determined by:
\begin{equation}
    \label{equ:spring_Damp}
    FC_{ij} = F^s_{ij} + F^d_{ij} = k(d_{ij} - d_s) + c\frac{d}{dt}(d_{ij} - d_s)
\end{equation}
where $k$ and $c$ are the spring stiffness constant and damping coefficient,
respectively. $d_{ij}$ is the distance between the robot $i$ and robot
$j$. $d_s$ is the length of the spring, which defines the safe communication
distance between the agents.

We use Equation~\ref{equ:simplified}, a simplified version of
Equation~\ref{equ:spring_Damp}, to generate a velocity control
law. The velocity of the robot is expressed in a similar way to a
spring force, while discarding the damping effect, which leads to a simpler
kinematic controller that is more compatible with the control strategy
described earlier in the robot's kinematic model. In fact, it is similar, yet
simpler, to gradient based velocity control~\cite{ji2007distributed}:

\begin{equation}
    u_i = k(d_{ij}-d_s)
    \label{equ:simplified}
\end{equation}
where $u_i$ is the velocity vector of the robot, $d_{ij}$ is distance between
robots $i$ and $j$, and $d_s$ is the desired communication distance between
the agents. We define $P_i$ to be the set of parents of robot
$i$ connecting $i$ to the root robot, either directly or through other robots
acting as communication relay.

\subsection{Bottom-up specification}
Robots in the swarm follow a simple set of rules using local information,
depending on their roles. There exists a parent-child relationship between the
robots in the chain.

\begin{algorithm}[ht]
  \caption{Worker robot control rules}\label{algo:worker}
  \begin{algorithmic}[1]
    \Procedure{worker routine}{}\label{algo:worker:proc}
    	\If {DISMANTLING}
    		\State $move\ in\ parents\ heading$
    		\If{root distance $\leq$ safe distance}
    		    \State $broadcast\ root\ dismantle\ complete$
    		\EndIf
    	\ElsIf {Parent unresponsive}
    		\State $move\ towards\ parent,\ find\ any\ robot\ in\ parents'\ link$
    		\If{root distance $\leq$ safe distance}
    		    \State $broadcast\ root\ dismantle\ complete$
    		\EndIf
    	\Else 
    	    \If{Parents in range and distance $\leq$ safe distance}
	 		    \State $Move\ towards\ target$
	 		\EndIf
	    	\If{Number of parents $\leq$ num of links required and distance  $\geq$ safe distance}
	    	    \State $Select\ a\ parent$ 
     		\EndIf
     	\EndIf
    \EndProcedure
  \end{algorithmic}
\end{algorithm}
\subsubsection{Workers}
A worker robot in the swarm initiates the growth of the chain, when it has reached the safe communication distance with the root. In that case, it chooses
new robots to act as relays. Once the selected robot/s are within a safe
communication distance, the worker starts to move towards the target. The
virtual spring and damper system ensures the integrity of the chain over
time. However, when a worker determines that one of its parents is unresponsive
using the method detailed in section~\ref{subsec:inter_agent}, it retracts the
chain by moving in the direction of the failed parent and reconnects the
broken link. The pseudo code listing~\ref{algo:worker} outlines the rules followed by
a worker to form the chain. The control input of
the worker robot $u_{i}^{w}$ is formulated as a sum of virtual forces:
\begin{equation}
    u_{i}^{w} 
    =
    \begin{cases}
      u_{i}^{sd} + f(d_{i}^{p})(u_{i}^{target} + u_{i}^{obstacle}), & \text{if}\ d_{i}^{p} < d_c, \forall p \in P_{i} \\
      u_{i}^{p}, & \text{otherwise}
    \end{cases}
\end{equation}
\begin{equation}
    \label{equ:worker_in_sum}
    u_{i}^{sd} 
    =
    \sum_{\forall j \in P} {u_{ij}^{sd}}
\end{equation}
where $u_{i}^{sd}$ is described in Equation~\ref{equ:simplified}, with
neighbours being all parents, as in
Equation~\ref{equ:worker_in_sum}. $u_{i}^{target}$ defines the control input to
attract a robot towards a target, and can be described as
  \begin{equation}
      u_{i}^{target} = k_{t}(p_i- p_t)
  \end{equation}
  where $p_i$ is the position of agent $i$, $p_t$ is the position of the
  target and $k_{t}$ is a constant gain.  $u_{i}^{obstacles}$ defines the
  control velocity that results from a repulsive potential created by
  obstacles, so as to let the robot move in a direction that avoids the
  obstacle, in a very similar way to what is described in
  \cite{shahriari2018lightweight}. $d_c$ is the
  critical communication distance above which communication becomes unreliable
  and results in a broken link. If the distance between the parent and the child
  increases above a critical communication distance $d_c$, the robot performs
  an emergency maneuver towards the parent using the virtual force created by
  $u_{i}^{p}$.
  \begin{equation}
      \label{equ:workercomfun}
      f(d_{i}^{p})
      = 
      \begin{cases}
      1, & \text{if}\ f(d_{i}^{p}) < d_s \\
      0, & otherwise
      \end{cases}
  \end{equation}

\begin{algorithm}[ht]
  \caption{Networker robot control rules}\label{algo:networker}
  \begin{algorithmic}[1]
    \Procedure{networker routine}{}\label{algo:networker:proc}
    	\If {child in view}
    		\If{new parent required to extend the chain}
    		    \State $Select\ new\ parent$
    		 \Else
    		    \State $Maintain\ safe\ distance\ between\ child\ and\ parent$
    		\EndIf
    	\ElsIf {new child not in view}
    	    \State $find\ child\ and\ move\ towards\ the\ child$
    	\ElsIf {old child not in view or unresponsive} 
	    	\State $move\ towards\ child\ and\ find\ any\ robot\ in\ chain$
	    \ElsIf {Parent unresponsive}
    		\State $move\ towards\ parent,\ find\ any\ robot\ in\ parents'\ link$
	    \ElsIf {DISMANTLING}
    		\State $move\ with\ parents\ heading$
     	\EndIf
    \EndProcedure
  \end{algorithmic}
\end{algorithm}
\subsubsection{Networker}
Networker robots act as communication relays, extending a chain in the
communication backbone for the worker robots to reach the
target. Algorithm~\ref{algo:networker} outlines the rules used by the
networker robots: when a networker gets selected to join a building chain, the
robot navigates to a safe distance from its child and maintains safe
communication distance and acts as a parent for the selecting robot. Moreover,
if the networker reaches a safe communication distance, if the chain needs to
be further extended, it selects a new robot to join. If a link in the chain
breaks in case of a robot failure, the robots at both ends of the broken links
take half of the responsibility to regain connection. The parent robots move
towards the children, and vice versa the children robot of failed robot move
towards the parent. If any of the robots of the same chain are encountered,
the connection is reestablished and the parent child relationship detailed above
starts with the newly bridged robots. The control law maintaining the
integrity of the networker position in the chain is:
\begin{equation}
    \label{equ:netcontrol}
    u_{i}^{n} 
    =
    \begin{cases}
      u_{i}^{p}, & \text{if}\ d_{i}^{p} >= d_c, \forall p \in P_i\\
      u_{i}^{sd} + f(d_{i}^{pc})(u_{i}^{obstacle}),  & \text{otherwise} 
    \end{cases}
\end{equation}
\begin{equation}
    \label{equ:net_sd__sum}
    u_{i}^{sd} 
    =
    u_{p}^{sd} + u_{c}^{sd}  
\end{equation}
where $u^{p}_i$ is the force that attracts a networker towards its parent, if the
distance is over the critical distance $d_c$, as in
Equation~\ref{equ:netcontrol}; $u_i^{sd}$ is the control law that ensures the
integrity of the networker position from its parent and child. If a networker
reaches a critical distance from its parent, its child takes the
responsibility of regaining safe communication distance. In other words, a
chain retracts if the communication distance gets above critical
distance. Equation~\ref{equ:netobstfun} enables and disables obstacle avoidance
if the distance between parent or child increases above $d_s$.
\begin{equation}
      \label{equ:netobstfun}
      f(d_{i}^{pc})
      = 
      \begin{cases}
      1, & \text{if}\ d_{i}^{p} < d_s\ \text{or}\ d_{i}^{c} < d_s \\
      0, & otherwise
      \end{cases}
\end{equation}

\begin{algorithm}[ht]
  \caption{Root and free robot control rules}\label{algo:rootfree}
  \begin{algorithmic}[1]
    \Procedure{Root routine}{}\label{algo:root:proc}
        \State $Listen\ to\ status\ broadcast\ from\ robots\ connecting\ a\ chain$
    	\If {Insufficient robots predicted}
    	    \State $Find\ chain\ with\ least\ robots\ and\ broadcast\ dismantle$
    	    \State $Find\ chains\ to\ expand$
    	    \State $Broadcast\ expand\ message\ to\ chains\ to\ expand$ 
     	\EndIf
    \EndProcedure
    \Procedure{Free routine}{}\label{algo:free:proc}
    	\If {New request received}
    		\State $Accept\ request$
    		\State $navigate\ to\ child$
    	\EndIf
    	\State $Compute\ LJ\ potential\ with\ root\ and\ other\ free\ robot$
    	\State $Use\ accumulated\ value\ as\ movement\ command$
    \EndProcedure
  \end{algorithmic}
\end{algorithm}
\subsubsection{Free and Root}
The role of the root robot in the swarm is to serve as a reference point to
build the communication backbone and monitor the growth of the chains. The root
listens to the broadcasts from the networkers to monitor
the expansion of the communication chains. If the root predict that there is an insufficient number of robots to build all the chains for all the targets, it
broadcasts messages to dismantle the chain with the least number of robots. The
pseudo code describing the root robot rules can be found in
Algorithm~\ref{algo:rootfree} at line~\ref{algo:root:proc}.
 
Free robots are the robots that get selected by the networkers or the workers
to act as relays for extending the chains. The free robots form a cluster
around the root using the force created by a Lennard-Jones (LJ)
potential~\cite{verlet1967computer}, which uses the position
of the root and the free robots to place the robots in a cluster around the
root's proximity. The control law defining the interaction between the free
robots and the root robot is:
\begin{equation}
    u_{i}^{lj} = \frac{\epsilon}{d_{ij}} \Bigg[ \Big[ \frac{\delta}{d_ij} \big]^{4} - \big[\frac{\delta}{d_{ij}} \big]^{2} \Bigg]
\end{equation}
\begin{equation}
    \label{equ:clfree}
    u_{i}^{f} = u_{i}^{lj} + u_{i}^{obstacle}
\end{equation}
The control law in equ.~\ref{equ:clfree} defines the control used by the free robots to maintain the cluster around the root over time. 

\subsection{Inter-agent information flow \label{subsec:inter_agent}}
The communication between the robots is gossip-based, with strictly local
broadcasts making information flow like in many insect colonies~\cite{dorigo2000ant}. We define four
different broadcast topics:
\begin{enumerate*}
    \item status broadcast;
    \item request and response;
    \item parent strand info; and
    \item child strand Info.
\end{enumerate*}
Every robot in the swarm locally broadcasts its current status under the
status topic, including its \textit{current role},
\textit{previous role}, \textit{parent need} and \textit{target chain}. Every robot
listens to the status message of all its neighbours. The status information is
serialized into a 4 byte value, with each sub-part of the status consuming 1
byte each. Each robot keeps lists for each of the roles, updated when receiving a new status message. From the updated lists, the list corresponding to
the free robots is used during the selection, process described
in~\ref{sec:approach} by a worker or networker. The
\textit{parent need} of the status broadcast is used by parent
robots to determine whether more extension to the chain is
required. The target robot broadcasts a status message with a \textit{parent
  need} of 1 if the target was not reached, and 0 otherwise. This
information flows all the way down to the robot connecting the root, and this
robot selects new parents if the target needs more expansion. Moreover, the
status messages are also used to predict robot failure. When a robot in a chain
determines the absence of a status message from a particular parent or a child
over a period of time, this robot is declared inactive. Inactive robots
result in broken links, a broken link is tackled using parent and child strand
info broadcasts. Using the strand info, the robot sensing broken link moves in the direction of the failed robot and tries to find any robot in the strand info that was connecting the failed robot and bridges connection with this robot that was after the failed robot. The robots while moving to bridge the connections of a failed link, moves in a way that ensures link connectivity with the rest of the chain by enforcing the forces described above.  

The parent strand info messages are a serialized string containing all the
parent robots in a chain up to the robot receiving the broadcast. This
broadcast starts from the root and flows through all the children in a
chain. The Root broadcasts its own id, which is received by its children, who
append their ID to the message and rebroadcast it, and so on, until it reaches
the worker robot. Using the parent info broadcast, a robot can determine the
chain of robots connecting it to the root. In case of an intermediate robot
failure, the robots can determine which robots to look for to reconnect the
chain. The child strand info messages are similar to the
parent strand info broadcasts, except that the information of the
chain flows from worker to root.

The request and response broadcast topic is used to send a request to a
neighboring robot, for instance to ask a free robot to join a chain. This
topic is also used by the robots to send a response to a request sent by a
robot.

\begin{figure}
    \centering
	\includegraphics[width=\columnwidth]{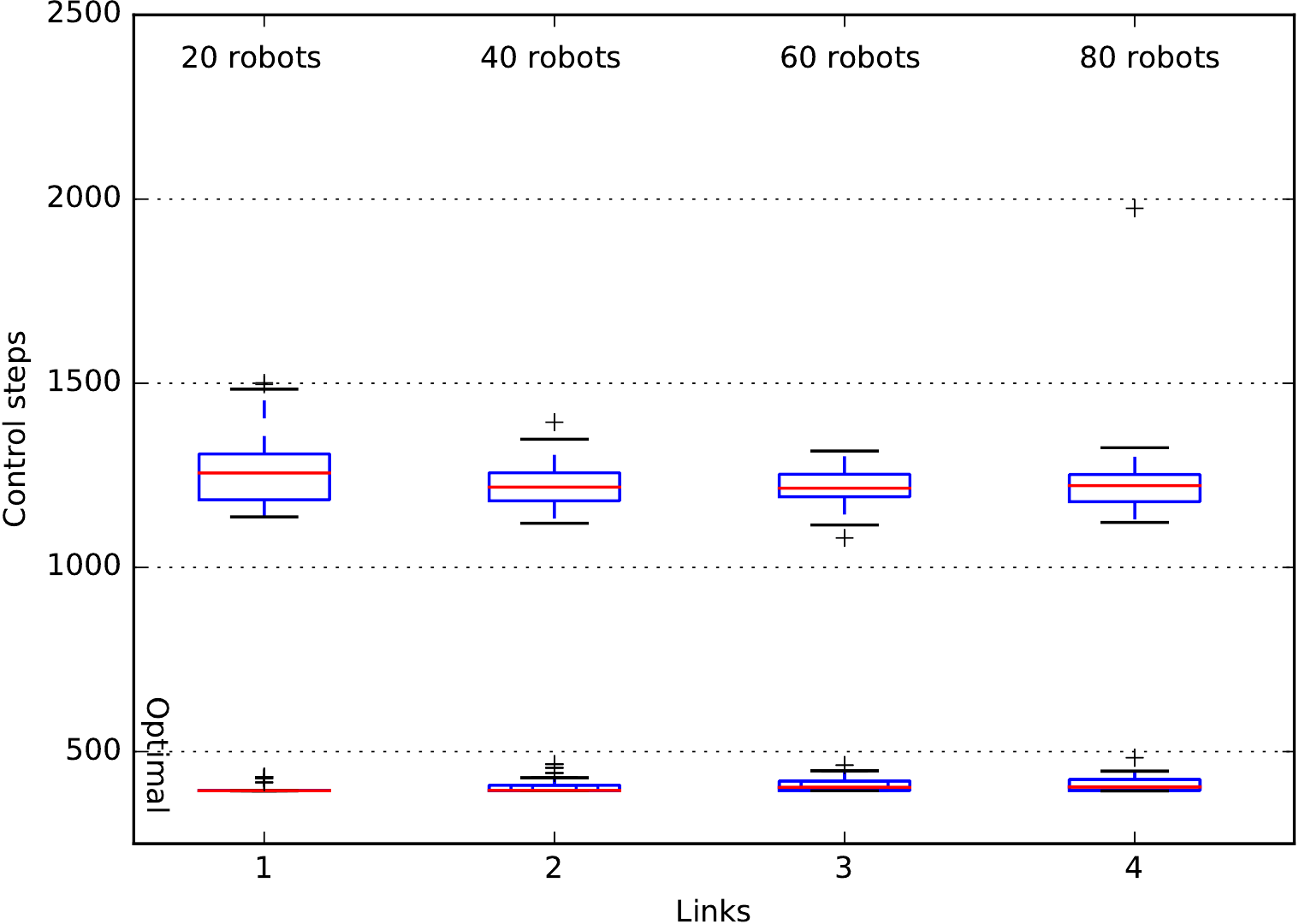}
	\caption{Time taken to build the chains with 20, 40, 60 and 80 robots.}
	\label{fig:time_to_completion}
\end{figure}

\section{Experiments~\label{sec:experiments}}
\subsection{Setup}
The experimental evaluations are aimed at studying the performance of the
algorithm under different conditions. The experiments were conducted using a
realistic physics-based multi-robot simulator
ARGoS~\cite{Pinciroli:SI2012}. The robot controllers are developed using
Buzz~\cite{PinciroliBuzz2016}, a programming language specific to swarm
robotics. We performed two sets of large scale simulations, using four
different link requirements. First we study the convergence properties of the
algorithm. In particular, Figure~\ref{fig:time_to_completion} reports the time needed by
the swarm to reach the targets while maintaining a connected network with the
desired number of links. This figure also reports the optimal time that might be required by the robots to navigate to a predefined location to form a topology, identical to the one formed by the proposed algorithm. The second set
of experiments observes different failure probabilities, and how they affect
our algorithm's performance. These injected failures make robots with an assigned backbone role
unresponsive (i.e., we remove their ability to communicate) with a random
probability, and report the time to reach targets.

\begin{table}[]
    \begin{center}
     \begin{tabular}{||c c c c||} 
     \hline
     Symbol & rationale & value & unit \\ [0.5ex] 
     \hline\hline
     $C$ & communication range & 2 & meters \\ 
     \hline
     $dt$ & simulation control step  & 0.1 & second  \\
     \hline
     $d_\delta$ & movement threshold & 0.3 & meters \\ 
     \hline
     $k$ & spring constant gain  & 0.8 & no unit \\ 
     \hline
     $d_s$ & safe communication distance & 1.4 & meters \\
     \hline
     $d_c$ & critical communication distance & 1.7  & meters  \\
     \hline
     $\epsilon$ & Lennard-jones potential epsilon  & 60 & no unit  \\
     \hline
     $\delta$ & Lennard-jones potential target & 0.50 & meters  \\ [1ex] 
     \hline
    \end{tabular}
    \end{center}
    \caption{Experimental design parameters used during the evaluations.}
    \label{tab:design}
\end{table}

In the first set of experiments, four different configurations were used, with
$N\in \{20,40,60,80\}$, where $N$ is the number of robots used in the
experiment. The second set of experiments involved robots with
$N\in \{40,80\}$. We placed the robots in a square 10x10 meters arena during
both experiments, the robots were given four targets, which are equally spaced
by 90 degrees from each other. The experimental design parameters used during
the evaluations are reported in table~\ref{tab:design}.

\subsection{Results}
Figure~\ref{Fig:argos_sim} reports the final configuration of the robot
network, with different configurations used during the evaluations. The
sub-figure, (a). shows the resulting formation of 20 robots reaching the targets
with a single link, (b). reports the 40 robot configuration with 2 links, and
(c). and (d). report the resulting communication chains with 60 (3 links)
and 80 robots (4 links) respectively.

\begin{figure}
    \centering
	\includegraphics[width=\columnwidth]{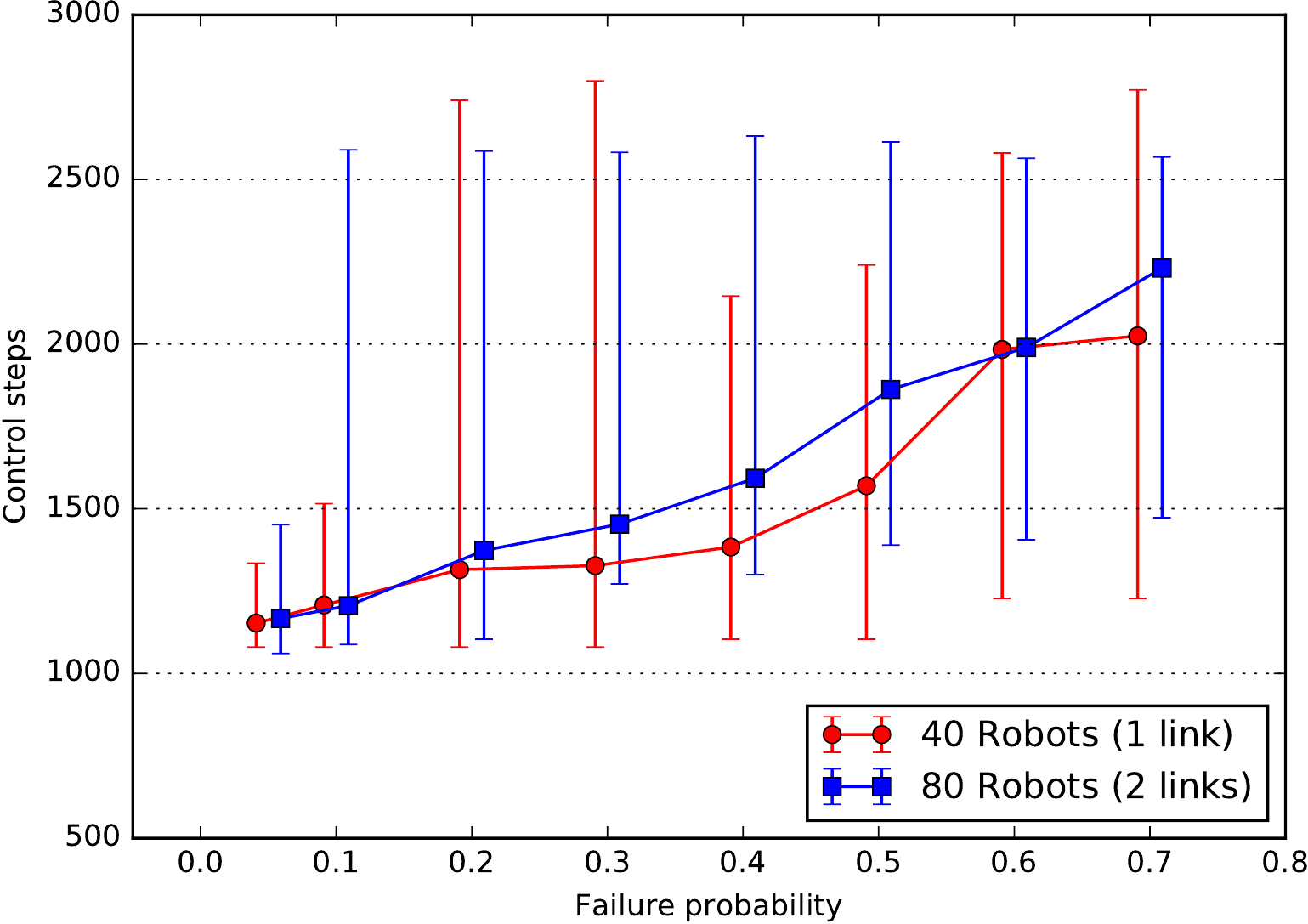}
	\caption{Time taken to build the chains with different percentage of fallible robots ( the markers are slightly offset for visual clarity).}
	\label{fig:time_to_completion_faults}
\end{figure}

\begin{figure}
    \centering
	\includegraphics[width=\columnwidth]{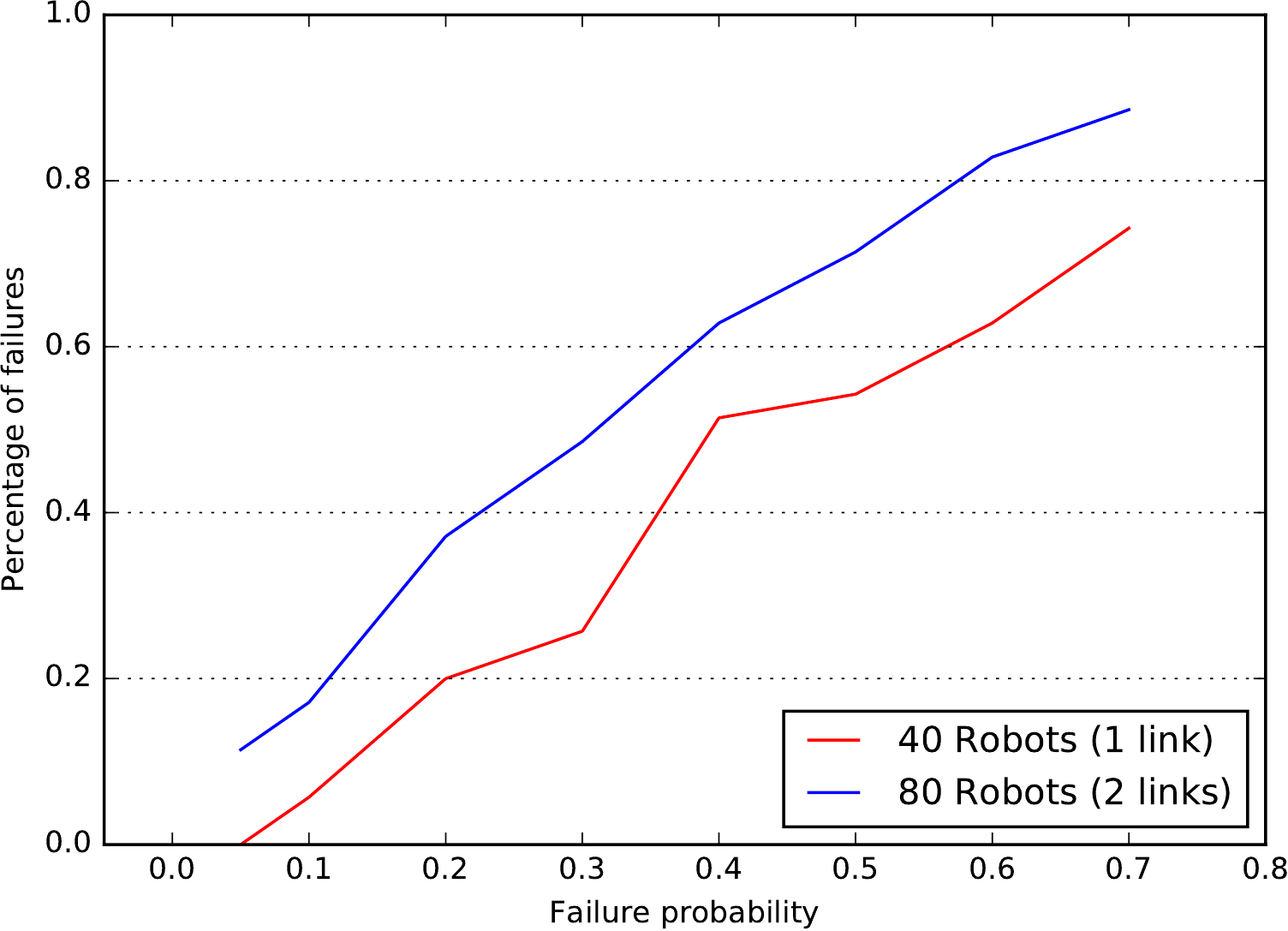}
	\caption{Percentage of mission failure vs robot faults.}
	\label{fig:failure_vs_faults}
\end{figure}

The box plot in figure~\ref{fig:time_to_completion} reports the number of time
steps required by the robots to build the chain and reach the distant
target locations over 35 trials, with each trial starting from random
locations. During the simulations, a single control step was set to
0.1s. Robots building a single-link chain consumed the highest time to
reach the targets, with a median of 35 trial correspond to 1400 control steps. The two-link and three-link configuration consumed about 1270 and 1223 control
steps. The four-link network with 80 robots interestingly consumed the least
control steps to reach the targets with 1214 median control
steps. 

The decreasing number of control steps with the increasing number of robots is
interesting, because opposite effect would be expected, if this were a centralized
or similar approach, the time consumed might increase as the number of robots
scales. The effect of decrease in the amount of time consumed with increase in
number links could be because of the increase in the summation of force that
is exerted on the robots with more links. Moreover, the optimal time consumed
by the robots in identical configurations is used as a baseline to indicate the performance of the algorithm. These optimal
time to reach the targets is computed, assuming the robot are in a perfect
world (no collisions, perfect knowledge of the control inputs to reach the
target).  In reality, neither these conditions is possible, nor the robots can
navigate without a control law.

\begin{figure}
    \centering
	\includegraphics[width=\columnwidth]{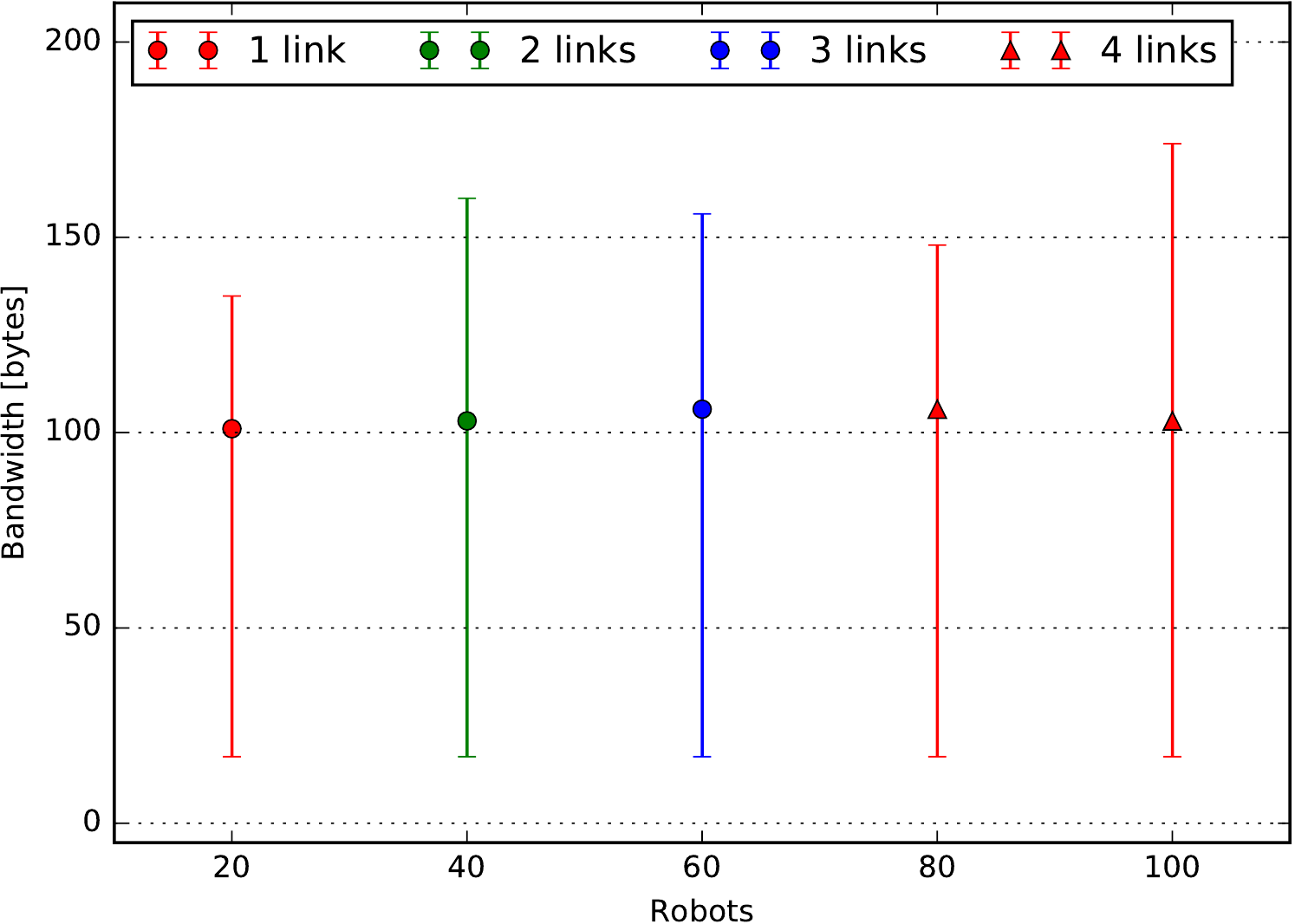}
	\caption{Max, min and median bandwidth consumed by any robot with the backbone construction approach over different configurations.}
	\label{fig:bandwidth}
\end{figure}

Figure~\ref{fig:bandwidth} reports the maximum, minimum, and median bandwidth
consumed by any given robot in the swarm during the simulations. The 10-robot
configuration consumed the minimal bandwidth of 17 bytes/timestep and the 100-robot case
consumed the maximum bandwidth of 174 bytes/timestep, however the median bandwidth were
around 100 bytes/timestep in all configurations. The major part of the bandwidth
consumed is contributed by the requests and responses sent by the robots to
form a chain. This class of messages depends on the average number of
neighbours for each robot. The other messages contributing to the bandwidth
consumption in increasing order are strand information broadcasts and status
messages.

\begin{figure}
    \centering
	\includegraphics[width=\columnwidth]{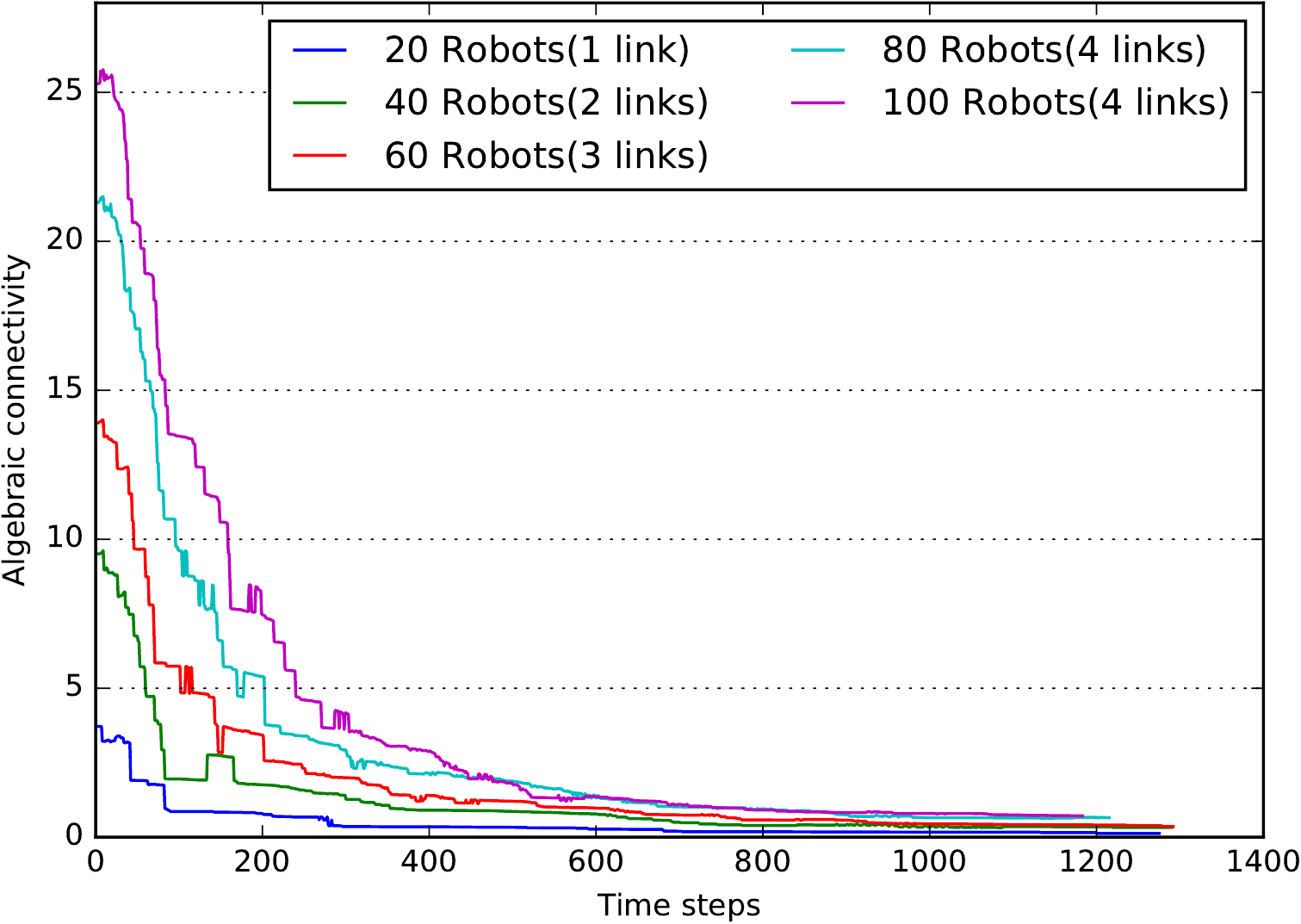}
	\caption{Evolution of algebraic connectivity over a trial of 5 different configurations.}
	\label{fig:algebraic}
\end{figure}

Figure~\ref{fig:algebraic} reports algebraic
connectivity ($\lambda_2$) of the resulting network graph. We observe that the
result is a very good representation of the underlying network
topology as introduced in section~\ref{sec:model}. During the initial stages of the simulation, the robots are in a
cluster (larger $\lambda_2$); as the experiment progresses $\lambda_2$ gets
close to zero, in particular with a single link network. This is expected,
since breaking a link  might result in a partition of the network.

Error bars in figure~\ref{fig:time_to_completion_faults} report the minimum,
maximum and median time consumed with different rates of failure over 35
trials. Intuitively, the time taken for the algorithm to converge increases as
the amount of fallible robots increase. The time consumed by the robots were
also quite fluctuating mainly due to the nature of the analysis. During this
analysis the 80 robot case with two links to maintain, consumed more time to
regain a chain break because of the robots trying to maintain two links at
once. The results follow a similar trend with up to 0.2 failure probability,
with almost identical medians. The time tread starts to diverge from 0.2
failure probability and climbs to 2067 and 2125 time steps at 0.7 failure
probability, for 40 and 80 robots respectively.
  Figure~\ref{fig:failure_vs_faults} illustrates the percentage of
mission failures with increase in faulty robots. 

\section{Conclusions \label{sec:conclusions}}
We proposed a decentralized approach to enforce connectivity constraints, capable of working alongside an existing algorithm given its low
computational and communication requirements. The algorithm progressively
builds a communication backbone for a set of robots visiting a distant
target. Our approach is self-organizing and inherently robust to single agent
failure. We tackle agent failures by propagating simple information through
the communication backbone.

We studied the performance of the proposed algorithm through a set of
simulation experiments that empirically demonstrate the
properties of the proposed algorithm through time to convergence, robustness
to failure and scalability to up to hundreds of agents. Our results show that
the algorithm withstands up to 70 percent agent failures and regains network
connectivity in the presence of broken communication links. Our approach
allows configuring the number of communication links either to increase
redundancy for critical missions or to provide more bandwidth for
communication hungry missions. Moreover, our approach provides methods to
tackle resource constrained scenarios -with fewer robots, by dismantling
un-grown chain's; constructing a single full chain, and visiting one target at
a time.  By design, the approach could tackle moving targets, however we did
not investigate this in this work, we plan to explore it in future works.

We envision to extend the approach in a number of ways, starting from
demonstrating its ability to run alongside other behaviours, to investigating
methods to tackle the availability of very few agents. We also plan to
investigate and deploy the algorithm on board a
KheperaIV~\footnote{https://www.k-team.com/mobile-robotics-products/khepera-iv}
ground robot and on a small swarm of flying UAV's (like DJI
m100s~\footnote{https://www.dji.com/matrice100} and 3DR
Solo~\footnote{https://3dr.com/solo-drone/}).


%
%
%

%
%
%
 \bibliographystyle{splncs04}
 \bibliography{aamas19}

\begin{thebibliography}{10}
\providecommand{\url}[1]{\texttt{#1}}
\providecommand{\urlprefix}{URL }
\providecommand{\doi}[1]{https://doi.org/#1}

\bibitem{aragues2014triggered}
Aragues, R., Sagues, C., Mezouar, Y.: Triggered minimum spanning tree for
  distributed coverage with connectivity maintenance. In: Control Conference
  (ECC), 2014 European. pp. 1881--1887. IEEE (2014)

\bibitem{Brambilla2013}
Brambilla, M., Ferrante, E., Birattari, M., Dorigo, M.: Swarm robotics: a
  review from the swarm engineering perspective. Swarm Intelligence
  \textbf{7}(1),  1--41 (Mar 2013). \doi{10.1007/s11721-012-0075-2},
  \url{https://doi.org/10.1007/s11721-012-0075-2}

\bibitem{cvetkovic2004}
Cvetkovic, D., Rowlinson, P.: Spectral graph theory. Topics in algebraic graph
  theory pp. 88--112 (2004)

\bibitem{de2006decentralized}
De~Gennaro, M.C., Jadbabaie, A.: Decentralized control of connectivity for
  multi-agent systems. In: Decision and Control, 2006 45th IEEE Conference on.
  pp. 3628--3633. IEEE (2006)

\bibitem{dorigo2000ant}
Dorigo, M., Bonabeau, E., Theraulaz, G.: Ant algorithms and stigmergy. Future
  Generation Computer Systems  \textbf{16}(8),  851--871 (2000)

\bibitem{fiedler1973}
Fiedler, M.: Algebraic connectivity of graphs. Czechoslovak mathematical
  journal  \textbf{23}(2),  298--305 (1973)

\bibitem{gasparri2017bounded}
Gasparri, A., Sabattini, L., Ulivi, G.: Bounded control law for global
  connectivity maintenance in cooperative multirobot systems. IEEE Transactions
  on Robotics  \textbf{33}(3),  700--717 (2017)

\bibitem{ghedini2016improving}
Ghedini, C., Ribeiro, C.H., Sabattini, L.: Improving the fault tolerance of
  multi-robot networks through a combined control law strategy. In: Resilient
  Networks Design and Modeling (RNDM), 2016 8th International Workshop on. pp.
  209--215. IEEE (2016)

\bibitem{Giuggioli2018}
Giuggioli, L., Arye, I., Heiblum~Robles, A., Kaminka, G.A.: From Ants to Birds:
  A Novel Bio-Inspired Approach to Online Area Coverage, pp. 31--43. Springer
  International Publishing, Cham (2018)

\bibitem{ji2007distributed}
Ji, M., Egerstedt, M.: Distributed coordination control of multiagent systems
  while preserving connectedness. IEEE Transactions on Robotics
  \textbf{23}(4),  693--703 (2007)

\bibitem{kim2005maximizing}
Kim, Y., Mesbahi, M.: On maximizing the second smallest eigenvalue of a
  state-dependent graph laplacian. In: American Control Conference, 2005.
  Proceedings of the 2005. pp. 99--103. IEEE (2005)

\bibitem{krupke2015distributed}
Krupke, D., Ernestus, M., Hemmer, M., Fekete, S.P.: Distributed cohesive
  control for robot swarms: Maintaining good connectivity in the presence of
  exterior forces. In: Intelligent Robots and Systems (IROS), 2015 IEEE/RSJ
  International Conference on. pp. 413--420. IEEE (2015)

\bibitem{Lee2018}
Lee, D.h.: {Resource-based task allocation for multi-robot systems}. Robotics
  and Autonomous Systems  \textbf{103},  151--161 (2018).
  \doi{S092188901730310X}

\bibitem{majcherczyk2018decentralized}
Majcherczyk, N., Jayabalan, A., Beltrame, G., Pinciroli, C.: Decentralized
  connectivity-preserving deployment of large-scale robot swarms. arXiv
  preprint arXiv:1806.00150  (2018)

\bibitem{manjanna2018heterogeneous}
Manjanna, S., Li, A.Q., Smith, R.N., Rekleitis, I., Dudek, G.: Heterogeneous
  multi-robot system for exploration and strategic water sampling. In: 2018
  IEEE International Conference on Robotics and Automation (ICRA). pp.~1--8.
  IEEE (2018)

\bibitem{minelli2018stop}
Minelli, M., Kaufmann, M., Panerati, J., Ghedini, C., Beltrame, G., Sabattini,
  L.: Stop, think, and roll: Online gain optimization for resilient multi-robot
  topologies. arXiv preprint arXiv:1809.07123  (2018)

\bibitem{panerati2018swarms}
Panerati, J., Gianoli, L., Pinciroli, C., Shabah, A., Nicolescu, G., Beltrame,
  G.: From swarms to stars: Task coverage in robot swarms with connectivity
  constraints. In: 2018 IEEE International Conference on Robotics and
  Automation (ICRA). pp. 7674--7681. IEEE (2018)

\bibitem{panerati2018robust}
Panerati, J., Minelli, M., Ghedini, C., Meyer, L., Kaufmann, M., Sabattini, L.,
  Beltrame, G.: Robust connectivity maintenance for fallible robots. Autonomous
  Robots pp. 1--19 (2018)

\bibitem{PinciroliBuzz2016}
Pinciroli, C., Beltrame, G.: Buzz: An extensible programming language for
  heterogeneous swarm robotics. In: International Conference on Intelligent
  Robots and Systems. pp. 3794--3800. IEEE (October 2016)

\bibitem{Pinciroli:SI2012}
Pinciroli, C., Trianni, V., O'Grady, R., Pini, G., Brutschy, A., Brambilla, M.,
  Mathews, N., Ferrante, E., {Di Caro}, G., Ducatelle, F., Birattari, M.,
  Gambardella, L.M., Dorigo, M.: {ARGoS}: a modular, parallel, multi-engine
  simulator for multi-robot systems. Swarm Intelligence  \textbf{6}(4),
  271--295 (2012)

\bibitem{sabattini2011decentralized}
Sabattini, L., Chopra, N., Secchi, C.: On decentralized connectivity
  maintenance for mobile robotic systems. In: Decision and Control and European
  Control Conference (CDC-ECC), 2011 50th IEEE Conference on. pp. 988--993.
  IEEE (2011)

\bibitem{sabattini2013decentralized}
Sabattini, L., Chopra, N., Secchi, C.: Decentralized connectivity maintenance
  for cooperative control of mobile robotic systems. The International Journal
  of Robotics Research  \textbf{32}(12),  1411--1423 (2013)

\bibitem{sahin2005}
{\c{S}}ahin, E.: Swarm robotics: From sources of inspiration to domains of
  application. In: {\c{S}}ahin, E., Spears, W.M. (eds.) Swarm Robotics. pp.
  10--20. Springer Berlin Heidelberg, Berlin, Heidelberg (2005)

\bibitem{Sampedro2018}
Sampedro, C., Rodriguez-Ramos, A., Bavle, H., Carrio, A., de~la Puente, P.,
  Campoy, P.: A fully-autonomous aerial robot for search and rescue
  applications in indoor environments using learning-based techniques. Journal
  of Intelligent {\&} Robotic Systems  (Jul 2018).
  \doi{10.1007/s10846-018-0898-1},
  \url{https://doi.org/10.1007/s10846-018-0898-1}

\bibitem{schuresko2009distributed}
Schuresko, M., Cort{\'e}s, J.: Distributed tree rearrangements for reachability
  and robust connectivity. In: International Workshop on Hybrid Systems:
  Computation and Control. pp. 470--474. Springer (2009)

\bibitem{schuresko2012distributed}
Schuresko, M., Cortes, J.: Distributed tree rearrangements for reachability and
  robust connectivity. SIAM Journal on Control and Optimization
  \textbf{50}(5),  2588--2620 (2012)

\bibitem{shahriari2018lightweight}
Shahriari, M., Svogor, I., St-Onge, D., Beltrame, G.: Lightweight collision
  avoidance for resource-constrained robots. In: Intelligent Robots and Systems
  (IROS), 2018 IEEE/RSJ International Conference on. IEEE (2018)

\bibitem{soleymani2015distributed}
Soleymani, T., Garone, E., Dorigo, M.: Distributed constrained connectivity
  control for proximity networks based on a receding horizon scheme. In:
  American Control Conference (ACC), 2015. pp. 1369--1374. IEEE (2015)

\bibitem{Stoy2001}
St{\o}y, K.: {Using situated communication in distributed autonomous mobile
  robots}. Proceedings of the 7th Scandinavian Conference on Artificial
  Intelligence pp. 44--52 (2001)

\bibitem{su2010rendezvous}
Su, H., Wang, X., Chen, G.: Rendezvous of multiple mobile agents with preserved
  network connectivity. Systems \& Control Letters  \textbf{59}(5),  313--322
  (2010)

\bibitem{verlet1967computer}
Verlet, L.: Computer" experiments" on classical fluids. i. thermodynamical
  properties of lennard-jones molecules. Physical review  \textbf{159}(1), ~98
  (1967)

\bibitem{Wiech2018}
Wiech, J., Eremeyev, V.A., Giorgio, I.: Virtual spring damper method for
  nonholonomic robotic swarm self-organization and leader following. Continuum
  Mechanics and Thermodynamics  \textbf{30}(5),  1091--1102 (Sep 2018).
  \doi{10.1007/s00161-018-0664-4},
  \url{https://doi.org/10.1007/s00161-018-0664-4}

\end{thebibliography}

\end{document}